\documentclass{article}
\usepackage{spconf,amsmath,graphicx}
\usepackage{hyperref}
\usepackage{cleveref}

\newcommand{\etal}{\textit{et al}. }
\newcommand{\ie}{\textit{i}.\textit{e}., }

\title{An End-to-end Framework For Integrated Pulmonary Nodule Detection and False Positive Reduction}
\name{Hao Tang$^{\dagger}$$^{\ddagger}$, Xingwei Liu$^{\ddagger}$, Xiaohui Xie\sthanks{Correspondence author.}$^{\dagger}$}
\address{$^{\dagger}$Department of Computer Science, University of California, Irvine, CA, USA \\
$^{\ddagger}$Deepvoxel Inc., 3200 Park Center Dr, Costa Mesa, CA, USA}
\begin{document}
%
\maketitle
\begin{abstract}
Pulmonary nodule detection using low-dose Computed Tomography (CT) is often the first step in lung disease screening and diagnosis. Recently, algorithms based on deep convolutional neural nets have shown great promise for automated nodule detection. Most of the existing deep learning nodule detection systems are constructed in two steps: a) nodule candidates screening and b) false positive reduction, using two different models trained separately. Although it is commonly adopted, the two-step approach not only imposes significant resource overhead on training two independent deep learning models, but also is sub-optimal because it prevents cross-talk between the two. In this work, we present an end-to-end framework for nodule detection, integrating nodule candidate screening and false positive reduction into one model, trained jointly. We demonstrate that the end-to-end system improves the performance by 3.88\% over the two-step approach, while at the same time reducing model complexity by one third and cutting inference time by 3.6 fold. Code will be made publicly available.
\end{abstract}

\section{INTRODUCTION}
\label{sec:intro}
Lung cancer has become the leading cause of cancer death among men and women worldwide \cite{siegel2015cancer}. Low-dose Computed Tomography (CT) has demonstrated to be an effective tool for detecting pulmonary nodules and screening lung cancer in early stages. Recent report suggests that detecting lung cancer in early stages can increase patients' 5-year survival rates by 63-75\% \cite{valente2016automatic}. However, locating nodules manually through CT scans is time-consuming. Over the past a few years, a lot of work has been done to automatically detect pulmonary nodules by using computer algorithms to read CT images. However, detecting pulmonary nodules with a low false positive rate while maintaining high sensitivity is challenging because of the variations in nodules' size, shape, and the abundance of tissues sharing similar appearance.

In recent years, deep convolutional neural nets have shown great promise for automated nodule detection \cite{dou2017multilevel,zhu2018deepem,tang2018automated,setio2017validation,DBLP:journals/corr/DingLHW17}. Most of the state-of-art nodule detection systems are constructed in two steps, composed of two separate subsystems: one used for generating nodule candidates, and the other for subsequent false positive reduction. The primary objective of the first subsystem is to generate a comprehensive list of candidate nodules with high sensitivity in mind, while the objective of the second subsystem is to remove false positives to improve specificity. Deep learning models have been proposed for both systems. The first subsystem usually uses segmentation-based methods or Region Proposal Network (RPN) \cite{DBLP:journals/corr/RenHG015} to generate candidates, while the second subsystem primarily uses classification models to distinguish nodules from non-nodules.

Although widely used, the two-step approach implemented in current deep learning systems has two major disadvantages. First, it is time-consuming and resource-intensive to construct and train two separate deep learning models. Although the objectives of the two subsystems are different, they share the commonality of extracting image features characterizing pulmonary nodules. As such, some of the model components can be shared and trained together. 
Second, the performance of the system may not be optimal because the two subsystems are trained separately without cross-talk between the two. 

Here we propose an end-to-end framework for pulmonary nodule detection, integrating nodule candidate generation and false positive reduction into a single model with shared feature extraction blocks, trained jointly. The new end-to-end system substantially reduces model complexity by eliminating one third of the parameters of the corresponding two-step model. It simplifies the training process and cuts the inference time by 3.6 fold. Experiments show that the end-to-end system also improves performance, increasing nodule detection accuracy by 3.88\% over the two-step approach. 


{\bf Related Work}
Deep learning, especially deep convolutional neural net (DCNN), has shown great success in medical image analysis. Ding \etal\cite{DBLP:journals/corr/DingLHW17} proposed a 2D regional proposal network for nodule candidate generation, followed by a 3D convolutional neural net for false positive reduction. Tang \etal \cite{tang2018automated} utilized 3D deep convolutional neural nets in both nodule candidate screening and false positive reduction. Zhu \etal \cite{zhu2018deepem} adopted 3D nodule candidate screening algorithm, and combined deep learning algorithm with a probabilistic model to explore the usage of weakly labeled clinical diagnosis data. There are also a few works focusing on false positive reduction, such as using multi-scale and model fusion to better classify nodules with various sizes \cite{dou2017multilevel} and using multi-view CNN for enhanced 3D information \cite{setio2016pulmonary}. Recent work also explored using single stage nodule detection model, for instance Khosravan \etal \cite{khosravan2018s4nd} proposed using single scale and single shot detection model, which however, has performance limitation because of its single scale assumption and the use of classification instead of detection when approaching this problem.


\section{PROPOSED METHOD}
\label{sec:method}

\begin{figure}
\centering
\includegraphics[width=0.45\textwidth]{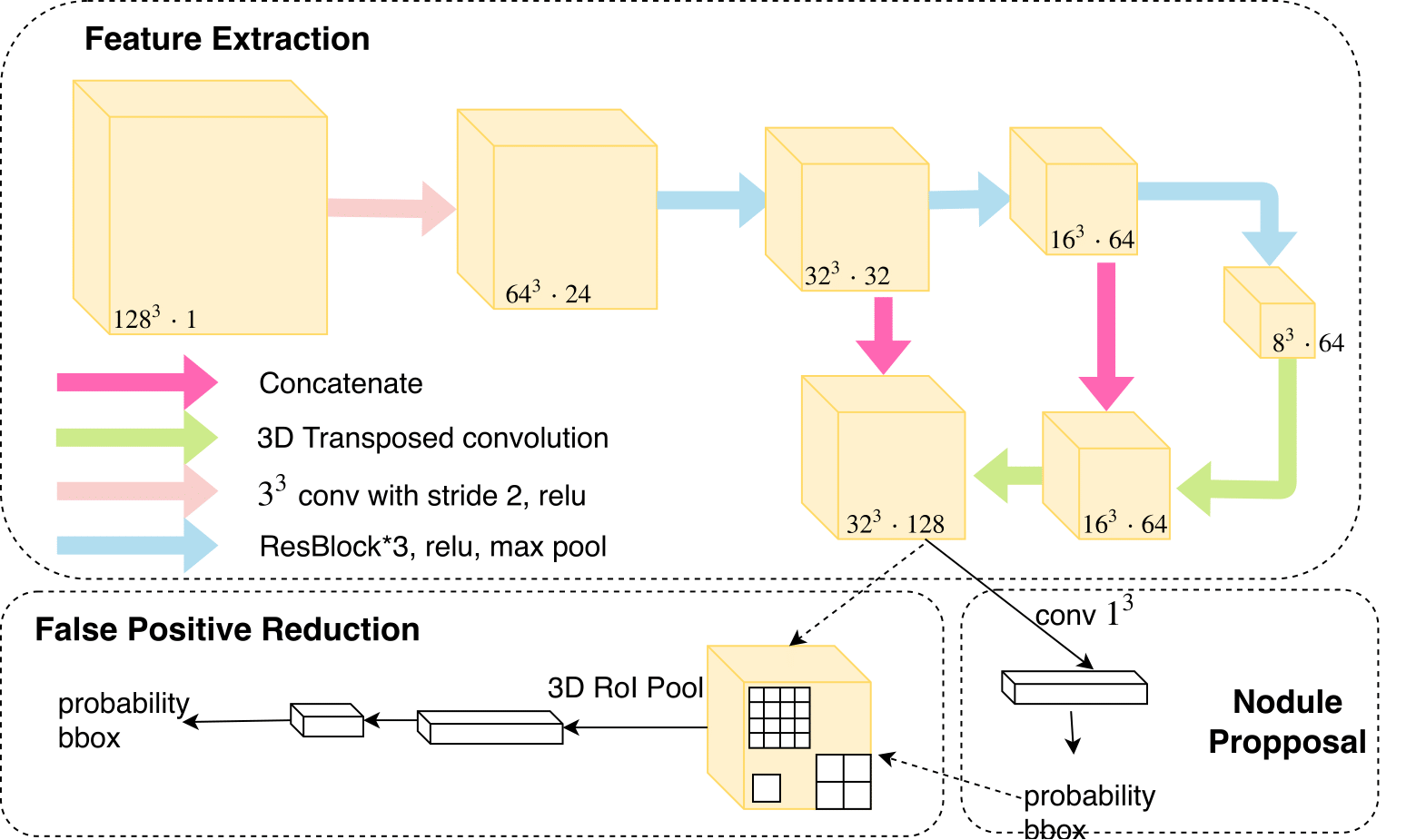}
\caption{End-to-end pulmonary nodule detection framework}
\label{fig:network}
\end{figure}

The proposed framework largely follows the two stages strategy: (1) generating nodule candidates using 3D Nodule Proposal Network, and (2) subsequent nodule candidate classification for false positive reduction. Different from the aforementioned works where two 3D DCNNs need to be trained separately, we discover the underlying computation of feature extraction for both networks can be shared and forwarded only once. Different tasks can be done on top of the feature map using different branches. The nodule candidate screening branch uses 3D Region Proposal Network adapted from Faster-RCNN \cite{DBLP:journals/corr/RenHG015}, and the predicted nodule proposal is then used to crop features of that nodule candidate using 3D Region of Interest (RoI) Pool layer, which are then fed as input to the nodule false positive reduction branch. The whole framework is shown in \Cref{fig:network}.

In feature extraction network, we use 3D convolution layer with stride 2 as the very first layer to reduce GPU memory cost. The subsequent convolution blocks are built using residual blocks \cite{DBLP:journals/corr/HeZRS15} with $3\times3\times3$ convolution followed by maxpooling to reduce spatial resolution. 

\subsection{Nodule Proposal Network}
The output of feature extraction is a $32\times32\times32$ feature map where each pixel on feature map has 128 feature channels. Then a $1\times1\times1$ convolution layer is applied to this feature map to generate $(z, y, x)$ coordinates, diameter and probability corresponding to the region of input volume. These five features are parameterized by five preset anchors of size 3, 5, 10, 20, 30. 

We compute a classification loss and four regression losses associated with $(z, y, x)$ and diameter for each of the anchor on each pixel on the feature map. We then use binary cross entropy loss with Online Hard-negative Example Mining (OHEM) for classification and $L1$ loss for four regressions.

Formally, our objective function is defined as:
\begin{equation}
L(\{p_i\}, \{t_i\}) = \frac{\sum_i{L_{cls}(p_i, p_i^*)}}{N_{cls}} + \lambda\frac{\sum_i{L_{reg}(t_i, t_i^*)}}{N_{reg}}
\end{equation}
where $i$ is the index of an anchor in one mini-batch and $p_i$ is the probability that anchor contains a nodule candidate. $p_i^*$ is 1 if an anchor is positive and 0 otherwise. $\lambda$ is a hyper-parameter for balancing classification and regression losses and we set it to 1 in this case. $N_{cls}$ is the total number of anchors considered for calculating the classification loss and $N_{reg}$ is the total number of anchors considered for calculating regression losses. $t_i$ is a vector representing the four parametrized coordinate offsets of the predicted bounded box and $t_i^*$ is the ground truth of the four regression terms. More specifically, $t_i=(t_z, t_y, t_x, t_d)$ is defined as:
\begin{equation}
\label{eq_loss}
\begin{split}
t=(\frac{z-z_a}{d_a}, \frac{y-y_a}{d_a}, \frac{x-x_a}{d_a}, \log\frac{d}{d_a}) \\
t^*=(\frac{z^*-z_a}{d_a}, \frac{y^*-y_a}{d_a}, \frac{x^*-x_a}{d_a}, \log\frac{d^*}{d_a})
\end{split}
\end{equation}
where $z, y, x, d$ denote square box's center coordinates and its diameter since we only need diameter to measure the size of a nodule. $z, z_a, z^*$ denote the predicted box, anchor box and ground truth box respectively (likewise for $y, x, d$).

\subsection{False Positive Reduction Network}
The bounding box regression terms are applied to each anchor, representing the actual spatial location and diameter of nodule candidate, which we call nodule proposal. We then use 3D RoI Pool operation to extract a small feature map from each RoI (\ie $4\times4\times4$). These features contain all the information about this nodule candidate and they go through two fully connected layers for predicting the probability that it is a nodule and four regression terms regarding its $(z, y, x)$ coordinates and diameter. 

We use nodule candidate whose probability is equal or greater than 0.5 for training this branch. A nodule candidate is considered as positive if it overlaps with a nodule with an intersection over union (IoU) larger than a threshold 0.5. In contrast, if it has an IoU less than 0.1 with a nodule, we consider it as negative. All other nodule candidates do not contribute to the classification loss and we only calculate regression losses for positive nodule candidates. Definitions of classification and regression losses are the same as \Cref{eq_loss}.

\subsection{Training}
We train the whole network in an end to end fashion. We first train the nodule proposal network using Stochastic Gradient Descent (SGD) for 60 epochs and then we train both network together for another 100 epochs. This is because, in the beginning the nodule proposal network predicts random nodule candidates which would be time-consuming for training the false positive reduction branch. Learning rate of SGD optimizer is scheduled as 0.01 initially, decreased to 0.001 after 80 epochs and 0.0001 after 120 epochs. 

To improve the generalization ability of the network, input volume is randomly shifted, randomly flipped along all 3 axis, and randomly scaled between 0.9 and 1.1.

\section{EXPERIMENTS AND RESULTS}
\label{sec:res}
We validated our framework on large-scaled Tianchi competition dataset\footnote{\url{https://tianchi.aliyun.com/competition/rankingList.htm?raceId=231601&season=0}}. It contains 800 CT scans from 800 patients with released ground truth label. The CT scans were annotated in a similar way to LUNA16 \cite{setio2017validation} with exact nodule location and diameter information. We used 600 CT scans for training and validation and another holdout 200 CT scans for reporting the performance of our model.

Free-Response Receiver Operating Characteristic (FROC) \cite{kundel2008receiver} analysis was adopted to quantify trade-off between sensitivity and specificity. We used the same evaluation metric as the LUNA16 challenge \cite{valente2016automatic} and the evaluation was performed by measuring the detection sensitivity and false positives per scan (FPs/scan). A nodule detection is considered positive if and only if its predicted location falls within a distance $R$ from the ground truth nodule's center, where $R$ is one half of nodule's diameter. The final Competition Performance Metric (CPM) is defined as the average sensitivity at seven predefined FPs/scan rates: 1/8, 1/4, 1/2, 1,2,4, 8.


\begin{figure}
\centering
 \includegraphics[width=3.3in]{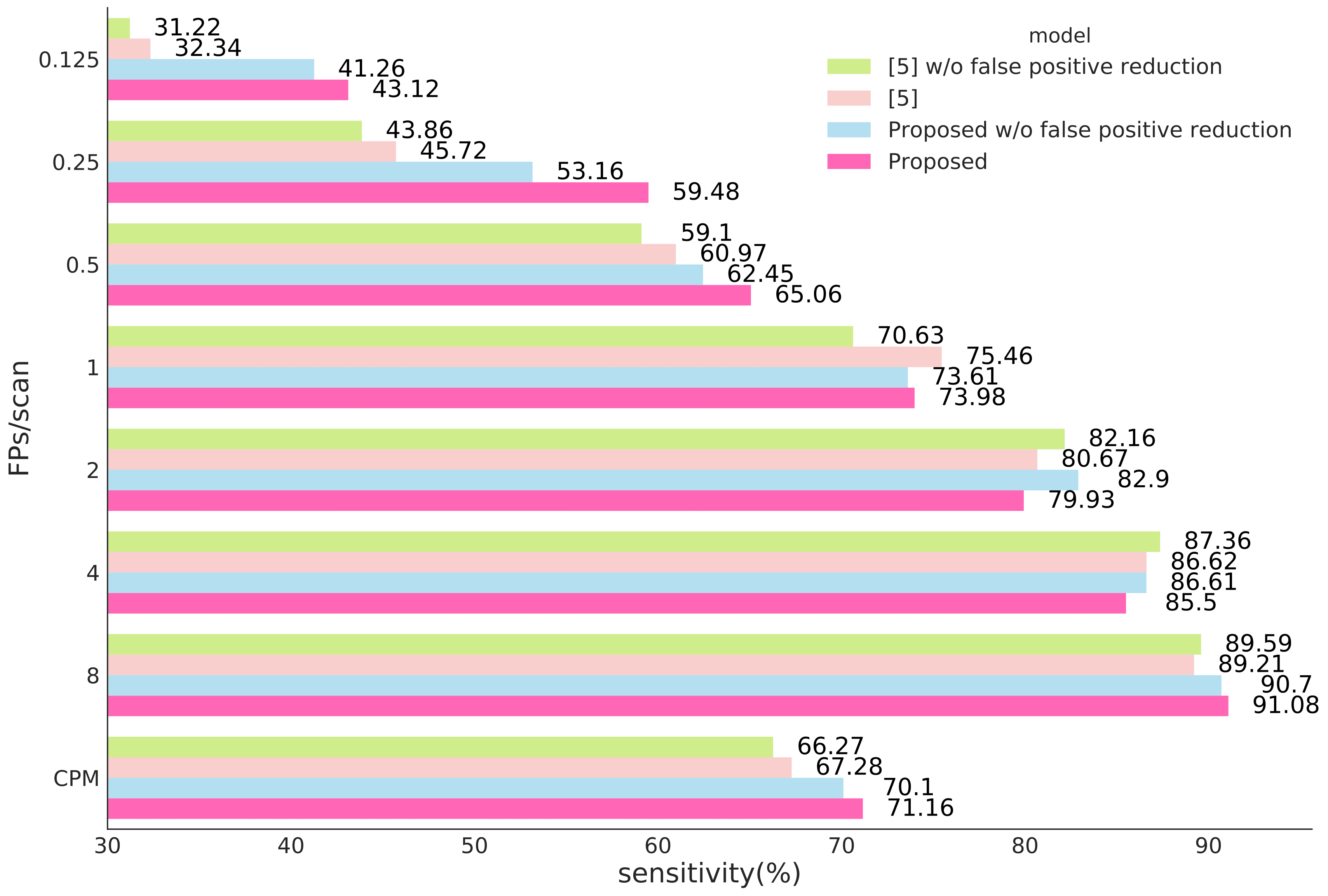}
\caption{Performance comparison}
\label{fig:performance}
\end{figure}

\begin{table}[]
\centering
\begin{tabular}{l r r}
\hline
\hline
                           & \# Parameters & Inference Time \\ \hline
\cite{tang2018automated}          &     $15903490$        &      $10.2 s$/CT            \\ \hline
Proposed      &      $9618523$         &       $2.8 s$/CT             \\ \hline

\end{tabular}
\caption{Comparison of number of parameters and time for inference between separate two stage framework \cite{tang2018automated} and the proposed framework}
\label{table:params}
\end{table}

\subsection{Performance comparison on holdout test set}
We compared performance among single stage nodule detection framework (\cite{tang2018automated} w/o false positive reduction), a state-of-art separate two-stage framework \cite{tang2018automated} and the proposed end-to-end two-stage framework. The step-wise gains of using the end-to-end framework is summarized in \Cref{fig:performance}. As we can see, when training the nodule proposal network and false positive reduction network together, the proposed end-to-end framework not only improves nodule proposal performance by 3.73\%, but further boosts the performance by 1.06\% using false positive reduction, which yields a 3.88\% improvement on CPM compared to previous state-of-art separate two-stage nodule detection model (\cite{tang2018automated}) without model ensemble.

Also, \Cref{table:params} shows the number of parameters used by the proposed framework, which is significantly lower than that of the previous two-stage model because of weight sharing. Moreover, since the proposed framework only needs to perform feature extraction once instead of forwarding the same patch of CT scan multiple times when inferring, it substantially reduces inference time for each CT scan from an average of $10.2s$ to $2.8s$ using single GPU.

\subsection{Visualization}
\begin{figure}
\centering
\includegraphics[width=0.45\textwidth]{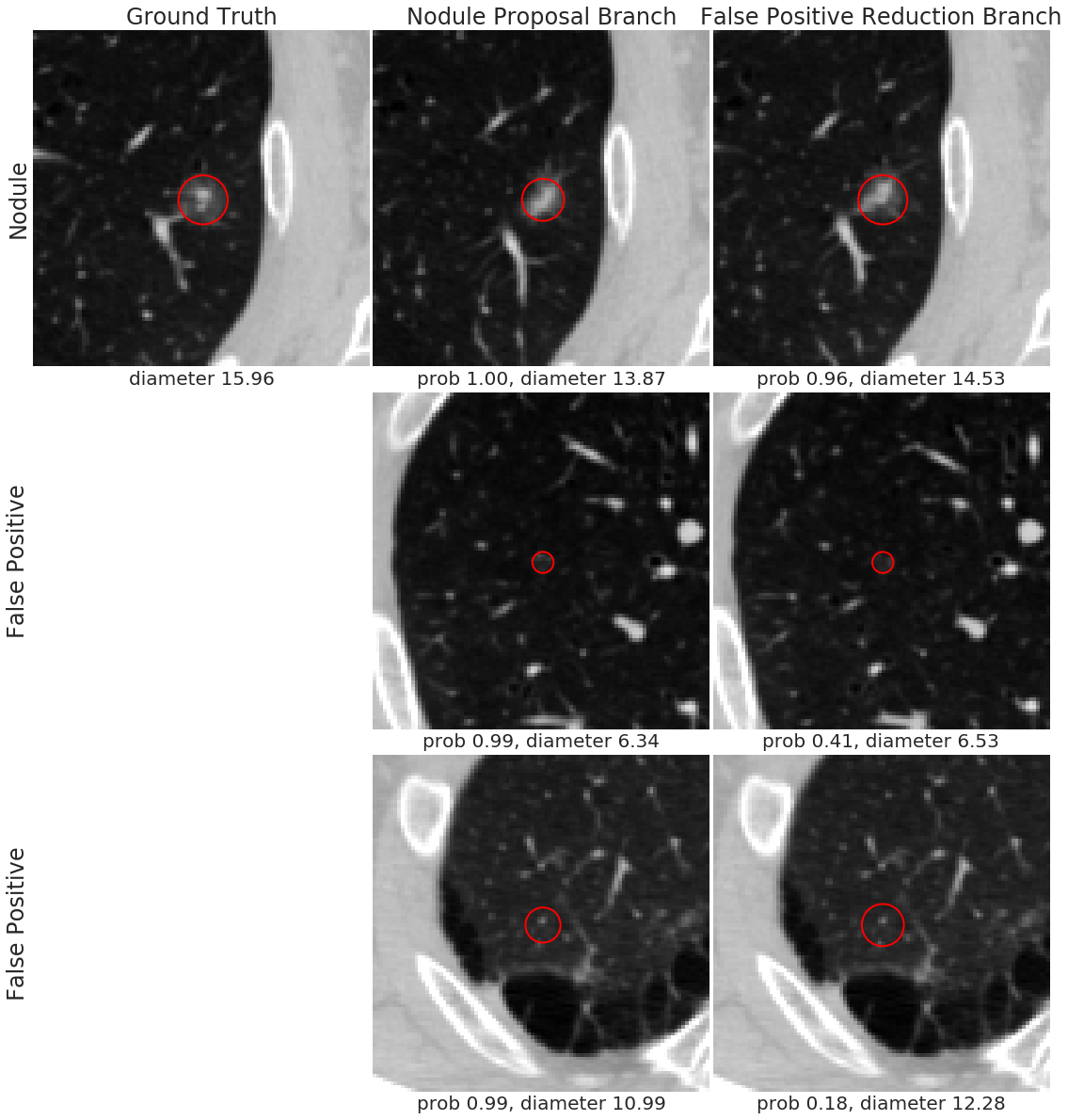}
\caption{Visualization of predictions from different branches of the proposed end-to-end nodule detection framework. The first row is from a true nodule while the other rows are false positives. We only show the center slice of each nodule candidate. Note that the false positive reduction branch is able to refine nodule diameter for the true nodule and significantly reduce probabilities for false positives.}
\label{fig:visualization}
\end{figure}

We randomly chose one patient from the holdout test set for visualizing performance gains of using the proposed framework in \Cref{fig:visualization}. The end-to-end model yields more precise detection of nodule location and size and better probability score, which demonstrates the proposed end-to-end framework improves the quality of pulmonary nodule detection.

\section{CONCLUSION}
\label{sec:conclusion}
In summary, we have presented a novel end-to-end framework for pulmonary nodule detection integrating nodule candidate generations and false positive reduction. The new system substantially reduces model complexity and inference time, thereby simplifying the training process and reducing resource overhead. Additionally, it improves the nodule detection performance over the two-step approach commonly used in existing nodule detection systems. Altogether, our work suggests that an end-to-end framework is more desirable for constructing deep learning-based pulmonary nodule detection systems.

\bibliographystyle{IEEEbib}
\bibliography{strings,refs}

\end{document}